\documentclass[11pt]{article}

\usepackage{acl}
\usepackage{times}
\usepackage{latexsym}
\usepackage{cuted}
\usepackage[utf8]{inputenc}
\usepackage[T1]{fontenc}
\usepackage{hyperref}
\usepackage{url}
\usepackage{amsfonts}
\usepackage{amsmath}
\usepackage{amssymb}
\usepackage{nicefrac}
\usepackage{microtype}
\usepackage{graphicx}
\usepackage{algorithm}
\usepackage{algorithmic}
\usepackage{multirow}
\usepackage{enumitem}
\usepackage{array}
\usepackage{longtable}

\usepackage{booktabs}
\usepackage{float}
\usepackage{tabularx}
\usepackage[table]{xcolor}
\usepackage{caption} 
\usepackage[breakable,skins]{tcolorbox}
\newtcolorbox{agentbox}[1][]{
  enhanced,
  breakable,
  colback=gray!6,
  colframe=gray!55,
  boxrule=0.4pt,
  arc=2pt,
  left=5pt, right=5pt, top=5pt, bottom=5pt,
  before skip=6pt, after skip=8pt,
  #1
}
\graphicspath{{paper/docx_media/media/}{docx_media/media/}}

\newcommand{\system}{\textsc{MAGG}}

\title{From Extraction to Governed Memory: Multi-Agent Knowledge Graph Construction with Domain-Expert Review}

\author{
\textbf{Pranav Bykampadi} \hfill \textbf{Neel Mokaria} \hfill \textbf{Vishesh Narayan} \\[0.5em] \textbf{Faizan Wajid} \hspace{3ex} \textbf{Ashok Agrawala} \\
  University of Maryland \\
  \texttt{\{pbykamp,nmokaria,vnaraya1,fwajid,agrawala\}@umd.edu}}

\begin{document}

\maketitle

\begin{abstract}

Knowledge graphs used by agentic systems are often treated as flat stores of extracted triples, with little record of who owns a fact, why it was admitted, or how it should be used downstream. We argue that reliable agentic knowledge systems require governance as an essential component of graph construction to bridge this gap. We propose \system{}, a principled multi-agent framework for constructing Governed Knowledge Graphs that introduces explicit governance decisions for reliable and trustworthy knowledge sharing. A domain classifier first induces entity and relation types directly from document content, enabling operation in open-world settings without fixed schemas. Candidate triples are assigned to domain owners, reviewed against supporting evidence, admitted through governance decisions, and stored with audit metadata. The same ownership structure is reused during question answering, where queries are routed to domain-specific graph experts rather than answered through undifferentiated retrieval. Our evaluation demonstrates \system{}'s effectiveness: On SciERC, \system{} improves strict triple F1 by 47\% and mapped triple F1 by 51\% over flat insertion. A blinded review of 120 triples finds governed-only triples more often source-supported than flat-only ones, and revised triples supported in 100\% of cases. Finally, on MuSiQue, \system{} outperforms Microsoft GraphRAG by 9.0 exact-match points and 11.2 token-F1 points.


\end{abstract}

\section{Introduction}

Knowledge graphs are increasingly used as memory substrates for agentic systems, but most current pipelines still treat them as flat stores of extracted triples. While other approaches have explored multi-agent coordination for knowledge graph construction \citep{lu2025karma, ye2024cooperkgc}, our system proposes a novel framework that integrates governance into the knowledge graph construction process from inception. This leaves several practical questions unanswered: 1) who owns a fact? 2) what evidence justified its inclusion? 3) how should conflicting updates be resolved and 4) how should downstream applications decide which part of the graph to consult? A flat graph can store relations, but it does not by itself provide the organizational structure needed for controlled updates, auditability, or domain-aware reasoning.

These limitations become more visible when knowledge graphs are used by active systems rather than passive analytics pipelines. Agentic systems must continually ingest new information, decide whether candidate facts should be admitted, revise facts when better evidence appears, and route queries to the right portion of the graph. In this setting, a graph is not merely a repository of extracted relations, but a memory structure that must support ownership, evidence-grounded review, incremental update, and downstream use. We argue that governance is therefore not an optional add-on, but a necessary graph layer for reliable agentic knowledge systems.

To this end, we developed \system{} as a knowledge representation system in which facts are admitted through explicit ownership, review, and provenance decisions. This organizational layer enables the same graph structure to support controlled construction, incremental updates, audit, and domain-routed reasoning. Furthermore, \system{}'s governance is exercised by LLM-based \textit{domain experts}: specialized agents that own domain-specific subgraphs, review candidate facts against evidence along with local graph context, and control admission into memory. We refer to these agents simply as Domain Expert Agents, which are LLM-based reviewers, not human experts.

\begin{figure}[t]
\centering
\includegraphics[width=0.45\textwidth]{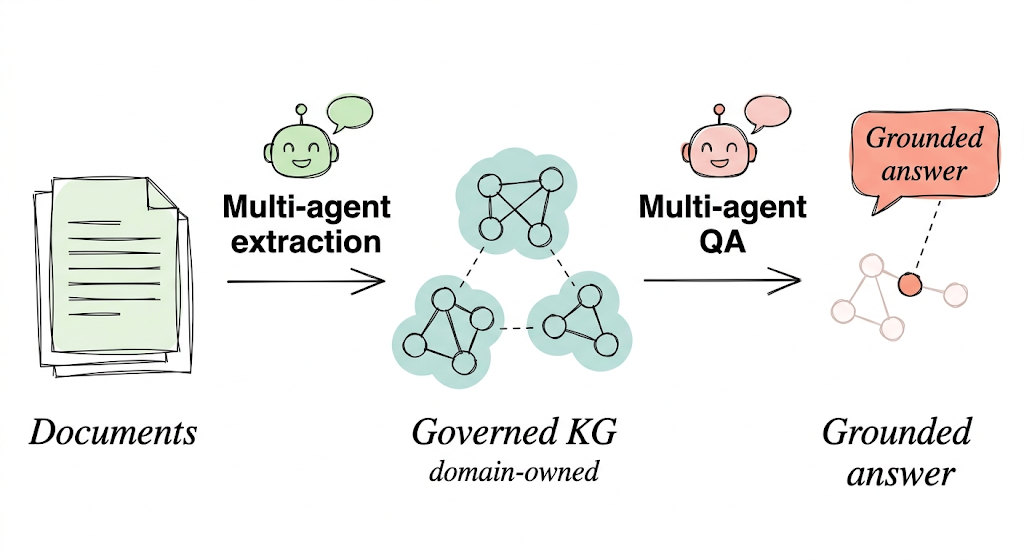}
\caption{Our framework parses documents into a governed knowledge graph through multi-agent deliberation and answers user queries through domain-expert collaboration over the resulting graph.}
\label{fig:overview}
\end{figure}

\system{} is organized into three coordinated layers. The \emph{creation layer} processes documents through a multi-stage extraction pipeline with blackboard-based deliberation for uncertain entities, relations, and triples. The \emph{governance layer} routes candidate triples to Domain Expert Agents, which review them against evidence and local domain memory and record each admission decision. The \emph{application layer} builds downstream functionality on top of the governed graph, including domain-routed question answering and incremental self-organization as new documents are processed. Figure~\ref{fig:overview} summarizes this flow, from raw documents through governed construction to domain-expert question answering. \textbf{Our central architectural idea is that governs what enters memory also determines how memory is later accessed and updated.}

We evaluate \system{} in both fixed-schema and open-world settings. These results support the broader claim of the paper: knowledge graphs for agentic systems should not be treated as flat collections of extracted triples, but as governed memory structures in which every fact has an owner, evidence, an admission path, and an audit trail. Experimental results are presented in \S\ref{sec:results}.

The contributions of this work are threefold. First, we formalize Governed Knowledge Graphs as a representation that augments entities and triples with domain ownership, explicit governance decisions, and LLM agent-mediated review. Second, we introduce \system{}, a multi-agent construction pipeline to operationalize these needs with governed admission and domain-routed Q/A. Third, we evaluate governance through extraction quality, governance ablations, human triple review, and downstream Q/A and show that the resulting domain-owned graph structure is useful for downstream graph-based question answering.

\section{Related Work}

\subsection{Knowledge Graphs \& Entity Extraction}
Knowledge graph construction has historically relied on supervised pipelines that
divided the extraction problem into sequential tasks: named entity recognition
(NER), relation classification, and coreference resolution. Early efforts to
unify these stages, such as the SciERC multi-task framework
\citep{luan2018multitask}, learned joint span representations across these
extraction tasks over scientific text. With the shift toward generative
architectures, these systems collapsed into single-pass operations.
\citet{cabot2021rebel} framed relation extraction as a seq2seq generation
problem covering over 200 relation types. Subsequent approaches, notably UIE
\citep{lu2022uie} and InstructUIE \citep{wang2023instructuie}, integrated all
core information extraction tasks under a single text-to-structure paradigm.
\citet{xue2024autore} modularized document-level extraction into distinct
generative phases for relation identification and endpoint binding, while
GLiNER \citep{zaratiana2024gliner} enabled zero-shot parallel span scoring for
arbitrary entity types without task-specific fine-tuning. These single-model
systems share a common failure mode: when faced with conflicting extractions,
hallucinated entities, or low-confidence boundary conditions, they have no
systematic mechanism to deliberate or resolve internal disagreement.

\subsection{Multi-Agent Systems}
To address the coordination and verification deficits of single models,
multi-agent frameworks have introduced agentic deliberation.
\citet{ye2024cooperkgc} demonstrated that a three-agent collaborative topology
significantly reduces extraction hallucinations compared to isolated models.
Scaling this approach, KARMA \citep{lu2025karma} deployed a nine-agent
architecture with specialized roles for schema alignment and conflict resolution
to enrich biomedical knowledge graphs. Concurrent research into multi-agent
deliberation topologies, including unstructured debate mechanisms \citep{du2024debate}, structured 
communication protocols \citep{chan2024chateval}, and iterative self-refinement 
\citep{madaan2023selfrefine} confirms that peer disagreement provides a more
reliable calibration signal than a single model's self-reported confidence
scores. Despite these architectural advances, the memory structures produced by
these systems remain fundamentally flat. They rely on a central controller to
merge verified extractions into a global store, stripping away metadata
regarding domain ownership, extraction provenance, and the governance rules
that authorized a triple's admission.

\subsection{QA Evaluation}
In the context of graph-grounded retrieval, GraphRAG \citep{edge2024graphrag}
constructs entity graphs and executes queries by map-reducing pre-computed
Leiden community summaries, imposing a purely topological hierarchy rather than
an ownership-based one. Similarly, while iterative graph-traversal techniques
\citep{sun2024tog,luo2024rog} outperform flat semantic retrieval for multi-hop
reasoning tasks, they treat the underlying knowledge graph as a static artifact
devoid of construction provenance. Evaluation frameworks have also shifted
toward granular verification. \citet{min2023factscore} popularized the
decomposition of generated text into atomic facts validated against retrieval
sources, a paradigm extended by SAFE \citep{wei2024safe} to achieve
human-annotation quality via web search. RefChecker \citep{hu2024refchecker}
sharpened the verification unit further to strict \textit{(subject, predicate,
object)} claim-triplets. LLM-as-a-judge approaches \citep{zheng2023judging}
scale evaluation cheaply but exhibit known positional and verbosity biases that
degrade reliability on structural claims.

None of these systems operate over a memory substrate with construction provenance. In a governed graph, the verdict on any claim should trace back to the domain owner and the governance decision that authorized its admission.

\section{Methodology}

\subsection{Governed Knowledge Graph}

Let $R$ be the relation vocabulary, $C \subseteq E \times R \times E$ the set of candidate triples, and $T \subseteq C$ the admitted triples. A Governed Knowledge Graph is $G=(E,C,T,D,\phi,\gamma,\pi,A)$, where $\phi:E\to2^D$ assigns entity ownership, $\gamma:C\to\Omega$ maps candidates to governance decisions, $\Omega=\{\texttt{approve},$ $\texttt{reject},$ $\texttt{revise},$ $\texttt{escalate},
$ $\texttt{auto\_approve}\}$, $\pi$ stores evidence/provenance, and $A$ is the audit log. A candidate $c$ enters $T$ only if $\gamma(c)\in\{\texttt{approve},$ $\texttt{revise},$ $\texttt{auto\_approve}\}$; escalations require later approval.

This definition differs from a standard knowledge graph in two ways. First, the graph contains an explicit organizational layer through domains and ownership. Second, admission into the triple set is mediated through governance rather than direct insertion. In implementation, this is realized through a two-phase protocol: propose a triple, obtain a governance decision, and then commit that decision. This changes the graph from a passive fact store into a managed memory structure.

The ownership function is written as $\phi: E \to 2^D$, rather than $\phi: E \to D$, because one entity may legitimately belong to more than one domain. In a scientific corpus, the same entity can participate in multiple conceptual regions of the graph: a method may belong to a methods domain and also sit inside a task-specific domain; a biomarker may belong to both an inflammation domain and a cardiometabolic domain. Using $2^D$ means $\phi(e)$ returns a set of owners, not a single owner. This is what makes cross-domain governance possible. Once entities have domain assignments, a triple can be classified as single-owner, cross-domain, or unowned depending on the ownership of its endpoints.

Operationally, the data structure consists of more than entities and triples. A governed triple also carries an ownership assignment, a governance action, a rationale, and an audit record. In the implementation, every admitted triple passes through \texttt{propose\_triple(...)} , produces a \texttt{GovernanceDecision}, and is then committed through \texttt{commit\_decision(...)}. This is the concrete mechanism that instantiates $\gamma$. The graph therefore stores not only what is known, but also how that knowledge entered memory and who was responsible for it.

\subsection{System Overview}

\begin{figure*}[t]
\centering
\includegraphics[width=0.8\textwidth]{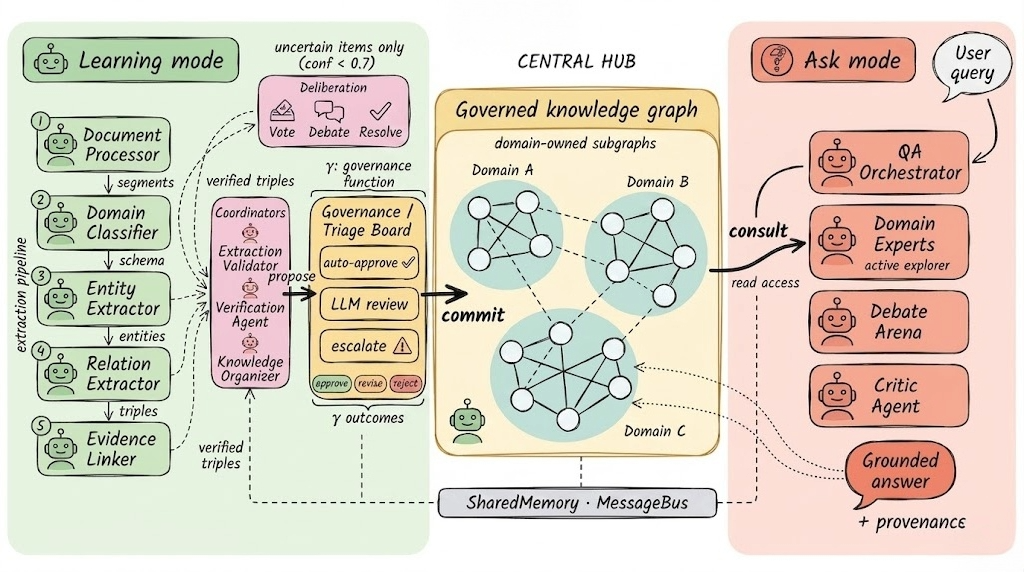}
\caption{System architecture. The framework operates in two modes over a shared governed knowledge graph. In \textbf{Learning mode} (left), worker agents run a sequential extraction pipeline, with uncertain items routed through a deliberation side-loop before governance review and commit. In \textbf{Ask mode} (right), a QA orchestrator routes user queries to domain experts that consult the graph through active exploration before returning a grounded answer with provenance.}
\label{fig:architecture}
\vspace{-10pt}
\end{figure*}

The system receives a corpus of documents. A top-level orchestrator manages the creation layer, which turns documents into entities and triples; the governance layer, which controls admission into memory; and the application layer, which uses the governed graph for downstream tasks. The pipeline is mostly sequential, and deliberation is activated only when an agent signals uncertainty or conflict.

The workflow is as follows; full construction-agent prompts and tool access are provided in Appendix~\ref{app:agents}.
\begin{enumerate}[leftmargin=*]
\item \texttt{DocumentProcessor:} segments each document into coherent chunks.
\item \texttt{DomainClassifier:} derives a working schema for the document. In benchmark mode it uses a fixed schema; in open-world mode it discovers entity and relation types from the content. From this schema, the system bootstraps preliminary domains before any triples are admitted. This is the first point at which governance becomes active during creation.
\item \texttt{EntityExtractor:} identifies entity mentions, canonical labels, candidate types, and domain assignments. The shortest-meaningful-noun-phrase rule is used only to normalize the entity span itself, so that entities are anchored to compact canonical labels rather than overly long surface forms. Relation extraction is not limited to noun pairs: downstream agents still use the full sentence context, including predicates, modifiers, and supporting evidence, to determine whether a typed relation should be proposed between extracted entities. The extractor also preserves short scientific terms, abbreviations, tool names, and task phrases rather than discarding them as generic text. These assignments partially instantiate the ownership function ($\phi$), allowing the system to know who owns a proposed fact during creation.
\item \texttt{RelationExtractor:} identifies relations between current entities and produces candidate triples. In fixed-schema benchmark mode, the extractor is constrained to the benchmark label set; in open-world mode, it can propose new relation structure.
\item \texttt{EvidenceLinker:} grounds each candidate triple in the source document by retrieving explicit supporting text spans and source references. It is instructed to quote or localize evidence directly from the document rather than generate free-form justifications, and these grounded spans are then inspected by the verification and governance stages before the triple can be admitted.
\item \texttt{VerificationAgent:} removes unsupported or low-quality extractions and outputs a verified set of triples suitable for admission review.
\item \texttt{KnowledgeOrganizer:} and governance layer take over. Each verified triple is proposed to the governed graph rather than inserted directly. The system routes the triple to the relevant domain or domains, produces a governance decision, commits the decision, and appends that decision to the audit log.
\end{enumerate}

Table~\ref{tab:agent_pipeline} summarizes the specialized agents used during construction. The pipeline is sequential by default: each agent transforms the current representation and passes it forward, while uncertain outputs are routed through deliberation before they can proceed to governance review.

\begin{table*}[h]
\centering
\footnotesize
\rowcolors{2}{gray!10}{white} 
\begin{tabularx}{\linewidth}{@{}p{0.22\linewidth}p{0.20\linewidth}p{0.22\linewidth}X@{}}
\toprule
\textbf{Agent} & \textbf{Input} & \textbf{Output} & \textbf{Purpose} \\
\midrule
DocumentProcessor & Documents & Chunks & Document segmentation. \\
DomainClassifier & Chunks & Working schema and domains & Schema/domain bootstrapping \\
EntityExtractor & Chunks & Entities & Mention \& canonical entity detection \\
RelationExtractor & Entities and context & Candidate triples & Typed relation proposal \\
EvidenceLinker & Triples and source text & Evidence spans & Provenance grounding \\
VerificationAgent & Triples and evidence & Verified candidates & Quality filter \\
KnowledgeOrganizer & Verified candidates & Governance decisions & Admission \& audit \\
\bottomrule
\end{tabularx}
\caption{Specialized agents in the MAGG construction pipeline.}
\label{tab:agent_pipeline}
\vspace{-10pt}
\end{table*}

Documents are processed incrementally: later documents can resolve entities against the existing graph, while new triples are still proposed from source evidence before being reviewed against prior domain memory. Thus, existing graph structure provides context for review but does not deterministically force new facts into old relation patterns.

\subsection{Deliberation, Blackboard, and Voting}

The pipeline is sequential by default; deliberation is an exceptional path triggered when an agent's confidence falls below a threshold of \textsc{0.7}, chosen to balance precision and compute efficiency, and serves as a middle ground between excessive automatic review and deliberation.

Rather than discarding all low-confidence items, the system holds them as \textit{hypotheses} for peer review, because many valuable facts are locally ambiguous rather than incorrect. Abbreviations, method names, and relation labels often require neighboring evidence to resolve. Deliberation lets the system repair uncertain candidates before governance review, rather than either admitting unstable triples or losing useful facts.

A hypothesis is a structured claim about an extraction unit that is not trusted enough for automatic continuation, typically an uncertain entity, triple, or relation  type, including its proposed content, posting agent, initial confidence, and supporting evidence. For example, from the sentence ``We present Minimum Bayes-Risk (MBR) decoding for statistical machine translation,'' the system may propose the candidate triple (\texttt{Minimum Bayes-Risk}, \texttt{Used-for}, \texttt{statistical machine translation}) at confidence 0.62. The subject span is under-specified relative to the source text, which refers more precisely to \texttt{Minimum Bayes-Risk decoding}, so the candidate is held in hypothesis state rather than admitted or discarded. This intermediate state is necessary because many low-confidence cases are not invalid, rather they are only locally ambiguous, and resolving them immediately preserves document context without polluting the graph with unstable structure.

The blackboard is a short-lived shared-memory workspace storing transient extraction state: pending hypotheses, vote requests, and vote rationales. We describe the shared memory architecture in more detail in Appendix~\ref{app:shared_memory}. A \textsc{DeliberationCoordinator} polls three role-conditioned voters--\textsc{EntityExtractor} (entity plausibility), \textsc{RelationExtractor} (relation coherence), and \textsc{EvidenceLinker} (textual support)--and aggregates a keep/no-keep decision. These differ from the Domain Expert Agents who govern admission; here, votes reflect extraction reliability rather than topical ownership. If votes conflict, the coordinator triggers a short structured debate. The final decision and rationale are written back to shared memory and used in two ways: operationally, as domain-local memory for later reviewers, and analytically, as evidence for debugging, ablation, and human validation. Accepted or revised candidates then proceed to governance review, where the governed graph stores only persistent, evidence-reviewed knowledge.

\subsection{Governed Graph Use}

\system{} relies on the governed knowledge graph as its foundation, providing a structured platform for downstream applications to build upon. We designed \system{} to support a range of applications and here we highlight two: question-answering, and incremental enrichment.

Downstream applications operate over the governed graph rather than over raw extracted text alone. For QA, an orchestrator reads the domain layout, routes questions to relevant domain experts, retrieves from their owned subgraphs and evidence, and synthesizes a final answer. For incremental enrichment, new documents are processed through the same ownership, evidence, and admission protocol, with graph updates under governed review.

\section{Results}
\label{sec:results}

We organize the results around four questions: whether governance improves admitted graph quality, 
whether that gain is explained by simple filtering, whether human review confirms that governed 
triples are better supported, and whether the governed graph improves downstream Q/A.

\subsection{Experimental Setup}

To answer the preceding questions, we evaluate \system{} across three empirical settings. The
first setting combines extraction quality and governance ablation because both ask how candidate
triples should enter memory.

\begin{enumerate}[leftmargin=*]
    \item \textbf{Extraction Quality and Governance Ablation:} We benchmark extraction performance against the SciERC test split and isolate the impact of governed admission over direct insertion, source-grounded insertion, and single-reviewer baselines. \system{} operates in a fixed-schema mode to compute exact precision/recall metrics (Entity F1, Strict/Mapped Triple F1) against a standardized target ontology.
    \item \textbf{Human Triple Review:} Because SciERC is a closed-world annotation, we conduct a blinded human adjudication study to evaluate actual source-support and graph utility for out-of-ontology triple proposals.
    \item \textbf{Downstream QA:} We test end-to-end multi-hop retrieval in an open-world setting using a 100-question slice of the MuSiQue dataset \citep{trivedi2022musique}, directly comparing our domain-routed QA approach against Microsoft GraphRAG.
\end{enumerate}

\subsection{Extraction Quality and Governance Ablation}

This subsection answers the first two results questions: whether governed admission improves the quality of the stored graph, and if the improvement is explained by simpler filtering. We evaluate all conditions on the full 100-document SciERC test split using GPT-5 and the fixed SciERC schema. The documents, target ontology, and base model are held fixed; only the admission pathway changes.

We compare four admission pathways. \textbf{Direct flat insertion} inserts extracted triples after basic structural cleanup, with no evidence linking, verification, domain ownership, or review. \textbf{Source-grounded flat insertion} adds EvidenceLinker and VerificationAgent, so triples are attached to source spans and checked for support before flat insertion. \textbf{Global reviewer} adds a single centralized LLM approve/reject reviewer, but no domain ownership or local graph memory. \textbf{\system{}} routes candidate triples to Domain Expert Agents, which review evidence and local graph context before approving, rejecting, or revising the triple for admission.

\begin{table*}[t]
    \centering
    \footnotesize
    \setlength{\tabcolsep}{5pt} 
    \rowcolors{2}{gray!10}{white}
    \begin{tabular}{lcccccccc}
    \toprule
    Condition & Entities & Triples & Strict P & Strict R & Strict F1 & Mapped P & Mapped R & Mapped F1 \\
    \midrule
    Direct flat insertion & 1742 & 2881 & 0.066 & 0.271 & 0.106 & 0.119 & 0.490 & 0.192 \\
    Source-grounded flat insertion & 1838 & 2912 & 0.067 & 0.277 & 0.107 & 0.123 & 0.510 & 0.198 \\
    Global reviewer & 1755 & 1599 & 0.103 & 0.234 & 0.143 & 0.194 & 0.444 & 0.271 \\
    \system{} & 1731 & 1077 & 0.129 & 0.199 & 0.156 & 0.240 & 0.369 & 0.290 \\
    \bottomrule
    \end{tabular}
    \caption{Extraction quality and governance ablation on the full 100-document SciERC test split. All rows use GPT-5 and the fixed SciERC schema. Mapped scores apply relation-label canonicalization and fuzzy entity normalization before scoring.}
    \label{tab:kg_results}
    \vspace{-8pt}
\end{table*}

We report strict and mapped triple precision, recall, and F1 in Table~\ref{tab:kg_results}. Strict scoring requires exact subject--relation--object agreement; mapped scoring applies relation canonicalization and fuzzy entity normalization before scoring. Precision captures how clean the admitted graph is, while recall captures how much of the benchmark graph is recovered. This distinction matters because direct insertion can improve recall by admitting many candidates, but at the cost of a noisier graph.

Source grounding alone does not explain the gain: source-grounded flat insertion barely moves strict F1 (0.106 → 0.107) while admitting nearly the same number of triples. A centralized global reviewer is stronger, reducing the graph to 1599 triples and improving strict F1 to 0.143, confirming that review helps but generic review is not the full mechanism.

\system{} performs best overall. It admits 1,077 triples, achieves the highest strict F1 (0.156) and mapped F1 (0.290), and has the strongest precision under both scoring regimes. Compared with direct flat insertion, strict precision rises from 0.066 to 0.129 and mapped precision rises from 0.119 to 0.240. The net effect is a cleaner admitted graph, with strict triple F1 improving by 47\% and mapped triple F1 by 51\% over direct GPT-5 insertion.

This supports the governance claim. The improvement is not reproduced by source grounding alone and is only partially reproduced by a single global reviewer. \system{} adds ownership, routing, domain-local memory, and subgraph-aware review, so Domain Expert Agents evaluate triples relative to the local graph context and prior accepted or rejected relation patterns. Because SciERC is a closed benchmark rather than an exhaustive fact inventory, some source-supported triples may still be counted as false positives; the human review study below complements this benchmark by checking whether such triples are actually supported.  

\subsection{Human Triple Review}

To assess whether governance changes the quality of admitted triples rather than merely changing graph structure, we conducted a lightweight human validation study over a blinded sample of 120 triples. The sample was randomly shuffled and drawn from four mutually exclusive categories: 40 triples admitted only by the governed graph, 40 triples admitted only by the FlatKG baseline, 20 triples admitted by both systems, and 20 triples revised by the governance layer before admission. During annotation, labels were hidden so that annotators did not know which of the four categories it came from.

\paragraph{Annotation instructions.}
Annotators were instructed to judge each triple using only the provided source evidence and not to rely on outside knowledge. For each triple, annotators evaluated four properties: whether the source evidence supported the triple, whether the relation label was correct, whether the subject and object endpoints were correct, and whether the triple was useful for graph-based question answering or retrieval. Source support was labeled as \emph{yes}, \emph{partial}, \emph{no}, or \emph{unclear}. Relation correctness and endpoint correctness were labeled as \emph{yes}, \emph{partial}, or \emph{no}. Usefulness was labeled as \emph{yes}, \emph{maybe}, or \emph{no}. A triple was considered useful if it encoded a relation likely to help graph search, multi-hop reasoning, domain routing, or source-grounded QA. We provide the exact prompt in Appendix \ref{app:anno_ins}.

\paragraph{Adjudication process.}
Two annotators judged whether each candidate triple was supported by the source evidence, whether its relation and endpoints were correct, and whether it was useful for graph-based QA. Disagreements were resolved by a third adjudicator. When labels differed, the adjudicated label reflected the most defensible interpretation of the evidence: direct textual support was required for a \emph{yes} label, plausible but incomplete or slightly under-specified support was marked \emph{partial}, and claims not grounded in the evidence were marked \emph{no}. We report both inter-annotator agreement (Table~\ref{tab:human_validation_agreement}) and the final adjudicated labels.

\begin{table}[h]
    \centering
    \small
    \begin{tabular}{lc}
        \toprule
        \textbf{Annotation Field} & \textbf{Agreement} \\
        \midrule
        Source support & 85\% \\
        Relation correctness & 75\% \\
        Endpoint correctness & 83\% \\
        KG/QA usefulness & 88\% \\
        \bottomrule
        \end{tabular}
    \caption{
    Inter-annotator agreement for the blinded human validation study. 
    Agreement was computed before adjudication over the 120 sampled triples.
    }
    \vspace{-8pt}
    \label{tab:human_validation_agreement}
\end{table}

The results support the governance hypothesis. Governed-only triples were judged supported or partially supported in 85.0\% of cases, compared with 62.5\% for FlatKG-only triples. They were also more often useful or potentially useful for QA, 77.5\% versus 62.5\%. Shared triples were similarly strong, with 85.0\% support and 85.0\% usefulness, suggesting that triples admitted by both systems form a high-confidence core. Most notably, revised triples were judged supported or partially supported in 100.0\% of cases and useful or potentially useful in 95.0\% of cases. Our results indicate that the governance layer does more than reject low-quality triples and emphasizes that the revision pathway is not merely rejecting bad triples; it repairs candidate facts into more supportable graph entries. We summarize the results of this experiment in Table~\ref{tab:human_validation_by_group}.
\begin{table*}[h]
\centering
\small
\begin{tabular}{lcccc}
\toprule
\textbf{Triple Group} 
& \textbf{$n$} 
& \textbf{Supported or Partial} 
& \textbf{Unsupported} 
& \textbf{Useful or Maybe Useful} \\
\midrule
Governed Only 
& 40 
& \textbf{85.0\% (34)} 
& 12.5\% (5) 
& \textbf{77.5\% (31)} \\

FlatKG Only 
& 40 
& 62.5\% (25) 
& 37.5\% (15) 
& 62.5\% (25) \\

Both 
& 20 
& 85.0\% (17) 
& 15.0\% (3) 
& 85.0\% (17) \\

Revised 
& 20 
& \textbf{100.0\% (20)} 
& \textbf{0.0\% (0)} 
& \textbf{95.0\% (19)} \\
\bottomrule
\end{tabular}
\caption{
Adjudicated human validation results over a blinded sample of 120 triples.
Governed-only triples were more often source-supported and useful for graph-based QA than FlatKG-only triples. 
Revised triples were judged supported or partially supported in all sampled cases, suggesting that governance can repair candidate triples before admission rather than merely filtering them. Note: percentages may not sum to 100 where adjudicated labels were marked unclear.
}
\label{tab:human_validation_by_group}
\vspace{-10pt}
\end{table*}

These findings complement the automatic SciERC evaluation. While benchmark triple matching penalizes any triple absent from the gold annotation, human validation shows that many governed-only triples are source-supported and useful for downstream graph reasoning. Governance therefore appears to improve the admitted graph not only by reducing unsupported triples, but also by shaping the graph toward more usable relational facts.

\subsection{Downstream QA on MuSiQue}

To evaluate whether the governed graph is useful for downstream applications, we test it on MuSiQue, a multi-hop question answering benchmark \citep{trivedi2022musique} designed so that answering a question requires combining facts across multiple pieces of evidence rather than retrieving a single local span. In contrast to the SciERC experiments, this setting is open-world: MuSiQue does not impose a fixed benchmark schema, so \system{} must organize entities, relations, and domains directly from unconstrained text.

We use a 100-question slice from MuSiQue 2-hop and report the results in Table~\ref{tab:musique}. In this setting, each question requires linking two supporting facts. For our experiment, the supporting-only corpus contains 200 contexts drawn from the question support sets. From this corpus, \system{} first constructs a governed graph in open-world mode by extracting entities and relations from the passages, organizing them into domain-owned subgraphs, and admitting facts through its governance pipeline. We then build a QA system on top of this graph: each question is decomposed and routed to the most relevant domain experts, those experts answer from their local subgraphs and neighboring evidence, and a final synthesis step combines their responses into a single answer with source verification. We compare this domain-routed QA system against official Microsoft GraphRAG and a flat KG-QA baseline. The Flat KG-QA condition uses the same QA orchestrator, query decomposition, and synthesis stack as \system{}, but operates over a flat-admitted graph with no domain ownership or governance review. The full QA agent roles and prompts are detailed in Appendix~\ref{app:qa_agents}.

\begin{table}[ht] 
\centering
\small
\begin{tabularx}{\columnwidth}{l@{\hspace{6pt}}c@{\hspace{6pt}}cX} 
\toprule
System & EM & F1 & Interpretation \\
\midrule
Flat KG-QA & 0.38 & 0.48 & Flat graph, same QA orchestrator \\
\addlinespace[3pt]
GraphRAG & 0.42 & 0.53 & Official graph-aware baseline \\
\addlinespace[3pt]
\system{} domain QA & \textbf{0.51} & \textbf{0.65} & Governed graph with domain-routed QA \\
\bottomrule
\end{tabularx}
\caption{MuSiQue QA, 100-Question Slice (GPT-5).}
\label{tab:musique}
\vspace{-10pt}
\end{table}

The 0.380 to 0.510 EM gain from Flat KG-QA to \system{} isolates the governance contribution: the QA stack is identical, while the underlying graph changes from flat-admitted to governed. The 0.420 to 0.510 EM gain over GraphRAG further shows that \system{}'s governed graph plus domain-routed QA outperforms a graph-aware baseline with its own retrieval approach. The same pattern holds for token F1, where \system{} reaches 0.646 compared with 0.480 for Flat KG-QA and 0.534 for GraphRAG. These results suggest that the ownership structure of the graph is useful at answer time: questions are routed through domain-owned subgraphs, domain experts return subgraph-grounded answers, and the final response is synthesized from these local views.

\section{Conclusion}

We introduced MAGG, a framework for constructing Governed Knowledge Graphs in which candidate triples are admitted through ownership, evidence review, governance decisions, and audit records rather than flat insertion. The same ownership structure is reused downstream for domain-routed question answering and incremental graph update.

Across SciERC, governance improves admitted triple quality over flat insertion, and ablations show that the gain is not explained by source filtering or a generic global reviewer alone. A blinded human review further suggests that governed and revised triples are more often source-supported and useful. On MuSiQue, MAGG improves over a graph-based QA baseline in a supporting-only multi-hop setting. These results support the central claim that governance is not merely a post-hoc filter, but a graph-level mechanism for organizing knowledge creation, update, audit, and downstream reasoning.

\newpage
\section*{Limitations}

MAGG improves admitted graph quality, but it does not solve extraction. Strict triple F1 remains modest on SciERC, reflecting the difficulty of exact triple matching and remaining errors in normalization, relation labeling, and evidence grounding. The system also introduces additional computational cost because domain-routed governance and domain-local memory require more LLM calls and prompt tokens than flat insertion. All experiments use GPT-5; whether governance gains transfer to other LLM families remains open. Finally, the open-world and human validation results are still preliminary: MuSiQue demonstrates open-world use, but we do not separately evaluate induced domain quality, and the blinded triple review should be expanded across more annotators, corpora, and domains.

\bibliography{main}

\clearpage
\appendix

\onecolumn 
\section{Token and Runtime Cost}
\label{app:cost}

We also report computational cost because \system{} trades additional governance calls for a cleaner and more auditable graph. Table \ref{tab:cost} reports exact API call and token counts from JSONL usage logs.

The current usage logs do not separate every internal extraction sub-agent into distinct cost buckets. Entity extraction, relation extraction, evidence linking, and any deliberation triggered inside extraction are included in the extraction-build usage log. Governance review is separately logged when replayed as a domain-memory review stage.

\begin{table*}[t]
\centering
\scriptsize
\setlength{\tabcolsep}{3pt} 
\renewcommand{\arraystretch}{0.95} 
\begin{tabular}{@{} l l rrrrrr @{}}
\toprule
Stage & Unit & LLM Calls & Prompt Tokens & Completion Tokens & Total Tokens & Calls / Unit & Tokens / Unit \\
\midrule
SciERC flat extraction build & 100 docs & 735 & 2,503,556 & 436,940 & 2,940,496 & 7.35 / doc & 29,405 / doc \\
SciERC source-grounded flat build & 100 docs & 801 & 2,490,259 & 655,658 & 3,145,917 & 8.01 / doc & 31,459 / doc \\
SciERC \system{} build & 100 docs & 1,845 & 6,707,359 & 1,313,298 & 8,020,657 & 18.45 / doc & 80,207 / doc \\
Domain-memory governance replay & 2912 candidate triples & 218 & 3,429,605 & 382,096 & 3,811,701 & 0.07 / candidate & 1,309 / candidate \\
Global reviewer replay & 2912 candidate triples & 145 & 238,524 & 119,151 & 357,675 & 0.05 / candidate & 123 / candidate \\
MuSiQue KG build & 200 supporting contexts & 630 & 6,599,190 & 829,591 & 7,428,781 & 3.15 / context & 37,144 / context \\
MuSiQue final \system{} QA & 100 questions & 105 & 255,689 & 145,612 & 401,301 & 1.05 / question & 4,013 / question \\
\bottomrule
\end{tabular}%
\caption{LLM Call and Token Cost}
\label{tab:cost}
\end{table*}

\section{Shared Memory}
\label{app:shared_memory}
\system{} uses shared memory to coordinate extraction within a document and to maintain consistency across documents. These functions are separated because the system must handle both short-lived local uncertainty and longer-lived accumulated knowledge. During extraction, agents need access to transient state such as pending hypotheses, intermediate votes, and active evidence spans. Across documents, the system needs persistent state such as admitted entities, aliases, domain ownership patterns, and prior governance decisions.

In practice, the memory system has four regions. \textbf{Working memory} stores the current pipeline state for the active document, including candidate entities, candidate triples, pending hypotheses, and vote requests. \textbf{Semantic memory} stores accepted graph structure and canonicalized entities. Once a candidate is revised and admitted, the canonical entity, its aliases, and the admitted triple are written to semantic memory for reuse in later extraction, governance, and QA. \textbf{Procedural memory} stores reusable review patterns derived from prior decisions, such as relation signatures that were previously accepted, rejected, or revised. \textbf{Episodic memory} stores per-document processing traces, such as what was proposed, revised, or rejected in a specific document.

These memory types are used at different timescales. Working memory is short-lived and is reused continuously during a single document pass. Semantic and procedural memory persist across documents and are consulted repeatedly during governance and downstream QA. Episodic memory is consulted less frequently, mainly when the system or a human needs to inspect how a prior decision was made. This separation is important because \system{} is not only extracting facts; it is building a governed memory that must distinguish temporary uncertainty from persistent knowledge.

\newpage
\section{Construction Agent Roles and Prompts}
\label{app:agents}

Each construction agent in \system{} is documented below using a
uniform schema (Inputs, Instruction/Prompt, Tools available, Output). Prompt text shown in italics is verbatim from the system prompt in our
implementation; longer prompts are summarized. Uncertain items at any
extraction stage are routed through the \texttt{DeliberationCoordinator}
(Appendix~\ref{app:deliberation_agent}) before continuing; the
coordinator is a side-protocol rather than one of the seven sequential
pipeline stages, but it is documented in its own subsection because it
is invoked by name throughout the workflow.

\subsection{Pipeline Sequence}

The construction pipeline executes the following order per document:
(1)~\textbf{DocumentProcessor}, (2)~\textbf{DomainClassifier},
(3)~\textbf{EntityExtractor}, (4)~\textbf{RelationExtractor},
(5)~\textbf{EvidenceLinker}, (6)~\textbf{VerificationAgent},
and (7)~\textbf{KnowledgeOrganizer}. The first agent receives raw
document text; the last writes admitted entities and triples into the
governed knowledge graph. The \textbf{DeliberationCoordinator} runs as
a side-loop between any agent that flags sub-threshold confidence and
the next pipeline stage.

\subsection{DocumentProcessor}
\label{app:doc_processor}

\begin{agentbox}
\textbf{Title:} \textit{DocumentProcessor}\\
\textbf{Role Description.} Entry point of the pipeline. Ingests raw
document text, segments it into semantically coherent overlapping
chunks, and registers the document in shared memory for cross-document
entity resolution.

\textbf{Inputs.}
\begin{itemize}[leftmargin=*,nosep]
\item \texttt{context: AgentContext} carrying \texttt{document\_id: str} and \texttt{text: str}
\item \texttt{source\_path: Optional[str]} containing the file path if available
\end{itemize}

\textbf{Instruction / Prompt.} No LLM call; segmentation is deterministic.
\begin{itemize}[leftmargin=*,nosep]
\item Unicode NFC normalization and control-character stripping.
\item Sentence-boundary split on \texttt{.!?}, then greedy packing into chunks of 1500--2000 characters with 150-character overlap (tail sentences re-added to the next chunk).
\item Each segment enriched with positional metadata (\texttt{segment\_id}, \texttt{char\_start/end}, \texttt{word\_count}, \texttt{is\_first}, \texttt{is\_last}).
\end{itemize}

\textbf{Tools available.}
\begin{itemize}[leftmargin=*,nosep]
\item \texttt{SharedMemory.} \texttt{register\_document} --- records the raw document for cross-document entity resolution.
\item \texttt{store\_in\_memory(EPISODIC, ...)} --- logs the processing event.
\item No LLM, no MessageBus, no deliberation.
\end{itemize}

\textbf{Output.} \texttt{ExtractionResult} with \texttt{items: List[Dict]} and \texttt{confidence: float} derived from segment-length quality and coverage heuristics.
\end{agentbox}

\newpage
\subsection{DomainClassifier}
\label{app:domain_classifier}

\begin{agentbox}
\textbf{Title:} \textit{DomainClassifier}\\
\textbf{Role Description.} Discovers the document's domain from
scratch (no predefined taxonomy) and generates a domain-specific
extraction schema---entity types, relation types, and few-shot
examples mined from the text. Broadcasts the schema to downstream
extraction agents.

\textbf{Inputs.}
\begin{itemize}[leftmargin=*,nosep]
\item \texttt{context: AgentContext}
\item \texttt{segments: List[Dict]} from \texttt{DocumentProcessor} (up to 4000 chars: first 2 + middle + last segments)
\end{itemize}

\textbf{Instruction / Prompt.} Two LLM calls, both asking the model to invent types rather than use predefined categories.
\begin{enumerate}[leftmargin=*,nosep]
\item \textbf{Domain analysis.} System: \textit{``You are an expert at discovering knowledge structures from scratch. NEVER use predefined schemas or standard taxonomies. Read the document carefully and INVENT a custom schema that fits THIS content. Types must be UPPER\_SNAKE\_CASE.''} Returns JSON with \texttt{primary\_domain}, \texttt{sub\_domains}, \texttt{entity\_types} \texttt{[\{type,} \texttt{description,} \texttt{priority,} \texttt{examples\_from\_text\}]}, \texttt{relation\_types} \texttt{[\{type,} \texttt{description,} \texttt{source\_types,} \texttt{target\_types,} \texttt{priority\}]}, \texttt{few\_shot\_examples}, and \texttt{extraction\_parameters} \texttt{\{complexity,} \texttt{knowledge\_density,} \texttt{requires\_coreference,} \texttt{has\_temporal\_relations,} \texttt{...\}}. Called with \texttt{self\_consistency(n\_samples=3,} \texttt{temperature=0.4)}.
\item \textbf{Relation examples.} Given the discovered types, extract one concrete example sentence with subject/object per relation type.
\end{enumerate}

\textbf{Tools available.}
\begin{itemize}[leftmargin=*,nosep]
\item LLM: \texttt{call\_llm}, \texttt{call\_llm\_} \texttt{with\_self\_consistency}.
\item \texttt{SharedMemory.} \texttt{store\_in\_memory(SEMANTIC, ...)} --- caches domain context.
\item \texttt{MessageBus.} \texttt{send\_message(INFORM, ...)} --- broadcasts entity types to \texttt{EntityExtractor} and relation types to \texttt{RelationExtractor}.
\item \texttt{escalate\_to\_coordinator} when confidence is below threshold.
\end{itemize}

\textbf{Output.} \texttt{ExtractionResult} with \texttt{items = [domain\_config]}. \texttt{context.domain} is set to \texttt{primary\_domain} for downstream agents. A generic fallback schema is returned if LLM parsing fails.
\end{agentbox}

\newpage
\subsection{EntityExtractor}
\label{app:entity_extractor}

\begin{agentbox}
\textbf{Title:} \textit{EntityExtractor}\\
\textbf{Role Description.} Four-stage entity extraction with
integrated coreference resolution. Discovers entity types beyond the
domain schema when needed, registers cross-document aliases, and
submits low-confidence entities to deliberation.

\textbf{Inputs.}
\begin{itemize}[leftmargin=*,nosep]
\item \texttt{context: AgentContext}
\item \texttt{segments: List[Dict]} from \texttt{DocumentProcessor}
\item \texttt{domain\_config: Dict} from \texttt{DomainClassifier}
\item Listens on \texttt{MessageBus} for \texttt{INFORM} messages carrying \texttt{entity\_types}.
\end{itemize}

\textbf{Instruction / Prompt.} Four sequential JSON-constrained stages.
\begin{enumerate}[leftmargin=*,nosep]
\item \textbf{Initial extraction.} \textit{``Extract entities based on what you observe in the text, not predefined categories. Err on the side of inclusion. Exclude generic dates, bare numbers, common adjectives.''} Output: \texttt{[\{text, start, end, type\_guess\}]}.
\item \textbf{Boundary refinement} (batches of 30). \textit{``You are an expert at identifying precise entity boundaries.''} Output: \texttt{[\{text,} \texttt{original\_text,} \texttt{start,} \texttt{end,} \texttt{boundary\_fixed\}]}.
\item \textbf{Type assignment} (batches of 25). Assigns discovered types in \texttt{UPPER\_SNAKE\_CASE}. Returns \texttt{type}, \texttt{type\_confidence}, \texttt{type\_reasoning}. Combined confidence = \texttt{(stage1\_conf + type\_conf + self\_consistency\_conf) / 3}.
\item \textbf{Coreference resolution} (batches of 20). Given the current batch and up to 20 known entities from KG/memory, groups mentions into canonical entities; aliases registered via \texttt{SharedMemory.} \texttt{register\_entity\_alias}.
\end{enumerate}
Self-consistency (\texttt{n\_samples=3}) is applied to stages 1 and 3 when enabled.

\textbf{Tools available.}
\begin{itemize}[leftmargin=*,nosep]
\item LLM: \texttt{call\_llm}, \texttt{call\_llm\_} \texttt{with\_self\_consistency}.
\item \texttt{SharedMemory.} \texttt{register\_entity\_alias}, \texttt{add\_entity\_context}, \texttt{store\_in\_memory(SEMANTIC)}.
\item \texttt{KnowledgeGraph} (read): loads up to 50 known entities for coreference.
\item \texttt{MessageBus.receive\_messages} (domain info).
\item Deliberation: \texttt{submit\_for\_deliberation(} \texttt{hypothesis\_type="entity",} \texttt{...)}; \texttt{evaluate\_hypothesis\_} \texttt{for\_vote} provides voting logic.
\item \texttt{escalate\_to\_coordinator}.
\end{itemize}

\textbf{Output.} \texttt{ExtractionResult} with \texttt{items: List[Dict]} carrying \texttt{text}, \texttt{type}, \texttt{confidence}, \texttt{aliases}, and span metadata.
\end{agentbox}

\newpage
\subsection{RelationExtractor}
\label{app:relation_extractor}

\begin{agentbox}
\textbf{Title:} \textit{RelationExtractor}\\
\textbf{Role Description.} Three-stage Relation--Head--First (RHF)
relation extractor. Identifies which relation types are present in
each segment, binds heads (subjects) for each relation, then resolves
tails (objects). Constrained to the benchmark label set in
fixed-schema mode; in open-world mode it can propose new relation
structure.

\textbf{Inputs.}
\begin{itemize}[leftmargin=*,nosep]
\item \texttt{context: AgentContext}
\item \texttt{segments: List[Dict]}, \texttt{entities: List[Dict]}, \texttt{domain\_config: Dict}
\item Listens on \texttt{MessageBus} for \texttt{INFORM} messages carrying \texttt{relation\_types}.
\end{itemize}

\textbf{Instruction / Prompt.} Three sequential LLM calls per segment, plus optional pairwise gleaning.
\begin{enumerate}[leftmargin=*,nosep]
\item \textbf{Stage 1 --- Relation identification.} Given the segment and entity list, return the subset of relation types from the schema that the segment plausibly expresses, each with a one-sentence rationale.
\item \textbf{Stage 2 --- Head binding.} For each identified relation, return \texttt{head\_bindings:} \texttt{[\{relation,} \texttt{head\_entity\_id,} \texttt{head\_text,} \texttt{evidence\_span\}]}. A short \emph{direction-only} hint is appended to the prompt to discourage subject/object inversion (e.g., for \texttt{used-for}, the head is the method being used).
\item \textbf{Stage 3 --- Tail binding} (batches of 30 head bindings). For each head binding, identify the OBJECT (tail) entity. \textit{``If you cannot find a valid, distinct object entity, SKIP that head binding entirely.''}
\end{enumerate}
Optional pairwise relation gleaning runs after Stage 3 for additional recall when enabled.

\textbf{Tools available.}
\begin{itemize}[leftmargin=*,nosep]
\item LLM: \texttt{call\_llm}.
\item \texttt{SharedMemory.} \texttt{store\_in\_memory(} \texttt{WORKING, ...)}.
\item \texttt{MessageBus.} \texttt{receive\_messages} (relation type schema).
\item Deliberation: \texttt{submit\_for\_deliberation(} \texttt{hypothesis\_type=} \texttt{"relation\_type"} \texttt{\textbar{} "triple", ...)}.
\item Schema enforcement: \texttt{\_enforce\_fixed\_} \texttt{schema\_relations} drops triples whose label is outside the fixed schema.
\end{itemize}

\textbf{Output.} \texttt{ExtractionResult} with \texttt{items} as candidate triples \texttt{\{subject,} \texttt{relation,} \texttt{object,} \texttt{confidence,} \texttt{evidence\_segment,} \texttt{stage\_diagnostics\}}. \texttt{metadata.} \texttt{relation\_funnel} records counts at each RHF stage for downstream diagnostics.
\end{agentbox}

\newpage
\subsection{EvidenceLinker}
\label{app:evidence_linker}

\begin{agentbox}
\textbf{Title:} \textit{EvidenceLinker}\\
\textbf{Role Description.} Attaches sentence-level evidence to each
triple, classifies evidence as explicit / implicit / inferred, and
optionally cross-references against prior KG knowledge to adjust
confidence. Submits weak-evidence triples to deliberation.

\textbf{Inputs.}
\begin{itemize}[leftmargin=*,nosep]
\item \texttt{context: AgentContext}
\item \texttt{triples: List[Dict]} from \texttt{RelationExtractor}
\item \texttt{segments: List[Dict]} from \texttt{DocumentProcessor}
\end{itemize}

\textbf{Instruction / Prompt.} Two LLM stages.
\begin{enumerate}[leftmargin=*,nosep]
\item \textbf{Evidence linking.} \textit{``Find supporting sentence(s). Label} \texttt{evidence\_type} \textit{$\in$ \{explicit, implicit, inferred\}. Set} \texttt{evidence\_strength} \textit{$\in [0,1]$. List any contradicting text with char positions.''} Batch size 3--20 (adaptive to token budget).
\item \textbf{Cross-reference} (optional, requires \texttt{enable\_cross\_reference}). \textit{``Compare new triples against up to 30 prior triples from KG/memory. Assign} \texttt{consistency\_status} \textit{$\in$ \{supported, contradicted, novel, refined\} and} \texttt{confidence\_adjustment} \textit{$\in [-0.3, +0.3]$.''}
\end{enumerate}
Final confidence is computed deterministically:
\begin{align*}
\text{final} &= \text{base\_confidence} \\
&\quad \times \text{evidence\_multiplier} \\
&\quad \times \text{evidence\_strength} \\
&\quad + \text{xref\_adjustment} \\
\text{evidence\_multiplier} &= \{\text{explicit}{:}\,1.0, \\
&\quad \ \text{implicit}{:}\,0.85, \\
&\quad \ \text{inferred}{:}\,0.7\} \\
\text{xref\_adjustment} &= \{\text{supported}{:}\,{+}0.15, \\
&\quad \ \text{contradicted}{:}\,{-}0.3, \\
&\quad \ \text{novel}{:}\,0, \\
&\quad \ \text{refined}{:}\,{+}0.1\}
\end{align*}

\textbf{Tools available.}
\begin{itemize}[leftmargin=*,nosep]
\item LLM: \texttt{call\_llm}.
\item \texttt{SharedMemory.} \texttt{store\_in\_memory(} \texttt{SEMANTIC,} \texttt{evidence\_linked\_triples=...)}, \texttt{retrieve\_from\_memory}.
\item \texttt{KnowledgeGraph} (read): pulls up to 50 prior triples.
\item Deliberation: \texttt{submit\_for\_deliberation(} \texttt{hypothesis\_type="triple")} for weak-evidence triples; \texttt{evaluate\_hypothesis\_} \texttt{for\_vote} provides evidence-based votes (\texttt{strong\_accept} for explicit+strong evidence, \texttt{reject} for contradictions).
\item \texttt{escalate\_to\_coordinator}.
\end{itemize}

\textbf{Output.} \texttt{ExtractionResult} with \texttt{items} carrying \texttt{evidence\_type}, \texttt{evidence\_strength}, \texttt{supporting\_text}, optional \texttt{consistency\_status}, and adjusted \texttt{confidence}.
\end{agentbox}

\newpage
\subsection{DeliberationCoordinator}
\label{app:deliberation_agent}

\begin{agentbox}
\textbf{Title:} \textit{DeliberationCoordinator} (protocol, not a \texttt{BaseAgent})\\
\textbf{Role Description.} Side-loop between extraction and validation.
Orchestrates multi-agent voting and structured debate over
entities and triples whose confidence falls in the uncertain band
$0.35 \le c < 0.7$. Items outside the band are accepted ($\ge 0.7$)
or rejected ($<0.35$) without a vote.

\textbf{Inputs.}
\begin{itemize}[leftmargin=*,nosep]
\item \texttt{entities: List[Dict]}, \texttt{triples: List[Dict]} from upstream agents
\item \texttt{segments: List[Dict]}, \texttt{context: AgentContext}
\end{itemize}

\textbf{Instruction / Protocol.} For each uncertain item:
\begin{enumerate}[leftmargin=*,nosep]
\item Submit a \texttt{Hypothesis(author,} \texttt{type$\in$\{entity,} \texttt{triple,} \texttt{relation\_type\},} \texttt{content,} \texttt{confidence,} \texttt{evidence)} to the blackboard.
\item Broadcast vote requests (\texttt{REQUEST}, \texttt{HIGH} priority) to the voting panel: \texttt{\{EntityExtractor,} \texttt{RelationExtractor,} \texttt{EvidenceLinker\}} by default.
\item Collect votes on a 7-level scale: \texttt{STRONG\_ACCEPT} ($+1.0$), \texttt{ACCEPT} ($+0.75$), \texttt{WEAK\_ACCEPT} ($+0.5$), \texttt{ABSTAIN} ($0$), \texttt{WEAK\_REJECT} ($-0.5$), \texttt{REJECT} ($-0.75$), \texttt{STRONG\_REJECT} ($-1.0$).
\item Weighted score $= \sum (\text{vote\_weight} \times \text{voter\_confidence} \times \text{agent\_weight})$ with \texttt{AGENT\_WEIGHTS} (\texttt{VerificationAgent}=1.5, \texttt{EvidenceLinker}=1.2, \texttt{EntityExtractor} = \texttt{RelationExtractor} = 1.0, \texttt{DomainClassifier}=0.8).
\item Debate is triggered if $|\text{weighted\_score}| < 0.3$: agents contribute support / oppose arguments over multiple rounds before final resolution.
\item Resolution: \texttt{min\_votes=2}, \texttt{consensus\_threshold=0.6}.
\end{enumerate}

\textbf{Tools available.}
\begin{itemize}[leftmargin=*,nosep]
\item \texttt{SharedMemory.} \texttt{post\_to\_blackboard}, \texttt{get\_blackboard\_entries}, \texttt{resolve\_blackboard\_entry}.
\item \texttt{MessageBus.send} (vote and debate requests), \texttt{receive}.
\item Per-agent delegation: \texttt{evaluate\_hypothesis\_} \texttt{for\_vote}, \texttt{provide\_debate\_argument}.
\end{itemize}

\textbf{Output.} Updated \texttt{entities} / \texttt{triples} lists containing items that survived voting or were high-confidence; counters \texttt{voting\_sessions}, \texttt{debates\_triggered}, \texttt{items\_accepted\_} \texttt{by\_vote}, \texttt{items\_rejected\_by\_vote}.
\end{agentbox}

\newpage
\subsection{VerificationAgent}
\label{app:verification_agent}

\begin{agentbox}
\textbf{Title:} \textit{VerificationAgent}\\
\textbf{Role Description.} Quality filter that turns the
evidence-linked candidate set into the verified set passed to the
\texttt{KnowledgeOrganizer}. The agent first runs an iterative
validate--refine loop to repair fixable issues in the candidate
entities and triples, then runs a final source-grounding verification
pass that drops unsupported or hallucinated triples and optionally
checks cross-document consistency against the existing graph. The
agent also processes blackboard escalations from upstream workers and
participates in debates orchestrated by the
\texttt{DeliberationCoordinator} (Appendix~\ref{app:deliberation_agent}).

\textbf{Inputs.}
\begin{itemize}[leftmargin=*,nosep]
\item \texttt{context: AgentContext} (carrying source text)
\item \texttt{entities: List[Dict]}, \texttt{triples: List[Dict]} from \texttt{EvidenceLinker} 
\item Blackboard escalations from any upstream worker
\item Listens on \texttt{MessageBus} for \texttt{DELEGATE} with \texttt{action="verify"}.
\end{itemize}

\textbf{Instruction / Prompt.} Three LLM stages plus deliberation participation.
\begin{enumerate}[leftmargin=*,nosep]
\item \textbf{Validation.} \textit{``You are an expert extraction validator.''} Given source text (up to 6K chars), entities, and triples, return per-item \texttt{\{valid,} \texttt{adjusted\_confidence,} \texttt{issues,} \texttt{corrections\}} plus overall \texttt{quality} $\in [0,1]$ and \texttt{recommendations}. Batch 40.
\item \textbf{Refinement.} \textit{``Apply corrections precisely.''} Emits \texttt{refined\_entities}, \texttt{refined\_triples}, \texttt{quality\_after\_refinement}. Batch 30. The validate--refine pair iterates while \texttt{overall\_quality} $<$ \texttt{quality\_threshold} and \texttt{iteration} $<$ \texttt{max\_refinement\_iterations} (default 4).
\item \textbf{Source verification.} \textit{``Accept BOTH explicit and reasonably inferred relationships. Only reject triples clearly contradicted or completely unsupported by the text. Partial support counts as valid with lower confidence.''} Per triple, return \texttt{verification\_status} $\in$ \{verified, partial, rejected, hallucinated\}, \texttt{final\_confidence}, \texttt{supporting\_evidence} (exact quote), and optional \texttt{rejection\_reason}. Batch 30. When prior KG triples exist, an additional cross-document consistency pass classifies each new triple as \{\texttt{consistent}, \texttt{contradicts}, \texttt{refines}, \texttt{redundant}\} and recommends \texttt{action} $\in$ \{\texttt{add}, \texttt{update}, \texttt{reject}, \texttt{merge}\}.
\item \textbf{Deliberation participation.} When invoked by the \texttt{DeliberationCoordinator}, the agent (a) maps \texttt{\{valid, confidence\}} to a \texttt{VoteType} (e.g.\ \texttt{valid=True, conf}$\ge$\texttt{0.8} $\rightarrow$ \texttt{STRONG\_ACCEPT}; \texttt{conf}$\le$\texttt{0.2} $\rightarrow$ \texttt{STRONG\_REJECT}) and (b) when prompted for a debate turn, returns \texttt{\{position, argument, key\_points\}} for the candidate under review.
\end{enumerate}
\textit{Decision logic:} \texttt{verified} $\rightarrow$ kept; \texttt{partial} $\rightarrow$ kept unless \texttt{strict\_mode=True}; \texttt{rejected} or \texttt{hallucinated} $\rightarrow$ dropped; remaining triples must satisfy \texttt{final\_confidence} $\ge$ \texttt{quality\_threshold} (0.45).

\textbf{Tools available.}
\begin{itemize}[leftmargin=*,nosep]
\item LLM: \texttt{call\_llm} for validation, refinement, and verification.
\item \texttt{SharedMemory.} \texttt{get\_blackboard\_entries(} \texttt{type="escalation")}, \texttt{resolve\_blackboard\_entry}, \texttt{store\_in\_memory(WORKING, ...)}.
\item \texttt{KnowledgeGraph} (read): up to 100 prior triples for cross-document consistency.
\item \texttt{MessageBus.receive\_messages} (\texttt{DELEGATE} escalations); \texttt{send DELEGATE} with \texttt{action="integrate"} to \texttt{KnowledgeOrganizer}.
\item Deliberation: \texttt{process\_pending\_} \texttt{deliberations}, \texttt{cast\_vote}, \texttt{provide\_debate\_argument}, \texttt{get\_deliberation\_results()}.
\item \texttt{DebugLogger.log\_decision} for every accept/reject.
\end{itemize}

\textbf{Output.} \texttt{ExtractionResult} where each approved triple carries \texttt{verification\_status}, \texttt{final\_confidence}, \texttt{supporting\_evidence}, optional \texttt{consistency\_status} and \texttt{consistency\_action}; \texttt{metadata.} \texttt{refinement\_iterations} reports validate--refine loops run, and \texttt{metadata.} \texttt{verification\_summary} reports \{total, verified, partial, rejected, hallucinated\} counts.
\end{agentbox}

\newpage
\subsection{KnowledgeOrganizer}
\label{app:knowledge_organizer}

\begin{agentbox}
\textbf{Title:} \textit{KnowledgeOrganizer}\\
\textbf{Role Description.} Final integration stage. Deduplicates
entities, normalizes relation names, resolves triple arguments to
canonical entity IDs, and writes everything into the shared
\texttt{KnowledgeGraph}. This is also the agent that proposes
candidate triples to the governance layer in governed mode.

\textbf{Inputs.}
\begin{itemize}[leftmargin=*,nosep]
\item \texttt{context: AgentContext}
\item \texttt{entities: List[Dict]} --- verified entities from \texttt{VerificationAgent}
\item \texttt{triples: List[Dict]} --- approved triples only
\item Listens on \texttt{MessageBus} for \texttt{DELEGATE} with \texttt{action="integrate"}.
\end{itemize}

\textbf{Instruction / Prompt.} Two LLM calls plus deterministic integration.
\begin{enumerate}[leftmargin=*,nosep]
\item \textbf{Entity deduplication.} (a)~Rule-based pass merges exact case-insensitive duplicates. (b)~LLM pass: \textit{``You are an expert at entity resolution.''} Returns \texttt{merge\_groups[} \texttt{\{canonical\_id,} \texttt{canonical\_name,} \texttt{merge\_ids,} \texttt{reason\}]}. Aliases registered in \texttt{SharedMemory.entity\_aliases}.
\item \textbf{Relation normalization.} \textit{``Normalize relation names consistently.''} Returns \texttt{normalizations[} \texttt{\{original,} \texttt{normalized,} \texttt{is\_inverse,} \texttt{reason\}]}. Cache stored in \texttt{self.relation\_mappings} to avoid re-querying.
\item \textbf{KG integration} (deterministic):
\begin{itemize}[leftmargin=*,nosep]
\item Filter garbage entities (empty, pure numeric, trivial pronouns like ``we'', ``it'', ``this study'', length $<2$).
\item Build \texttt{name\_to\_id} lookup from entity text, IDs, aliases, existing KG labels, and shared-memory aliases (prevents phantom entities).
\item Resolve subject/object strings to existing entity IDs; if unresolved, create a new entity of type \texttt{UNRESOLVED} rather than dropping the triple.
\item Call \texttt{KnowledgeGraph.} \texttt{add\_entity(...)} and \texttt{add\_triple(...)}.
\end{itemize}
\end{enumerate}

\textbf{Tools available.}
\begin{itemize}[leftmargin=*,nosep]
\item LLM: \texttt{call\_llm} for dedup and normalization.
\item \texttt{KnowledgeGraph} (write): \texttt{add\_entity}, \texttt{add\_triple}.
\item \texttt{SharedMemory.} \texttt{register\_entity\_alias}, \texttt{store\_in\_memory(SEMANTIC, ...)}.
\item \texttt{MessageBus.receive\_messages} for incoming \texttt{integrate} delegation.
\item Reporting: \texttt{get\_kg\_stats()}, \texttt{export\_knowledge\_graph()}.
\end{itemize}

\textbf{Output.} \texttt{ExtractionResult} with \texttt{metadata.kg\_stats} = \texttt{\{total\_entities,} \texttt{total\_triples,} \texttt{entity\_types,} \texttt{relation\_types,} \texttt{unique\_relations\}} and \texttt{metadata.integration\_stats} = \texttt{\{entities\_added,} \texttt{entities\_merged,} \texttt{triples\_added,} \texttt{triples\_updated,} \texttt{relations\_normalized\}}. \emph{Side effect:} the shared \texttt{KnowledgeGraph} is mutated and can be exported as JSON.
\end{agentbox}


\newpage
\section{QA Agent Roles and Prompts}
\label{app:qa_agents}

The application layer reuses the governed graph for downstrresizeboxeam
question answering. Three agents participate in each query:
the \textbf{QAOrchestrator}, which decomposes a user question and
routes it to one or more \textbf{DomainExpertAgents}; the
\textbf{DomainExpertAgents}, which answer from their owned subgraphs;
and a \textbf{SynthesisAgent}, which combines the per-domain answers
into a single grounded response. Each agent is documented below using
the same schema (Inputs, Instruction/Prompt, Tools available, Output)
as the construction agents in Appendix~\ref{app:agents}.

\subsection{QAOrchestrator}
\label{app:qa_orchestrator}

\begin{agentbox}
\textbf{Title:} \textit{QAOrchestrator}\\
\textbf{Role Description.} Top-level coordinator of question answering.
Reads the domain layout of the governed graph, decomposes the user
question into sub-questions, routes each sub-question to one or more
domain experts, gathers cross-domain bridge context, and invokes
synthesis.

\textbf{Inputs.}
\begin{itemize}[leftmargin=*,nosep]
\item \texttt{question: str}
\item \texttt{governed\_kg} --- the persistent governed graph and its domain layout
\end{itemize}

\textbf{Instruction / Prompt.} Single LLM call for decomposition and routing.
\begin{enumerate}[leftmargin=*,nosep]
\item \textbf{Decomposition and routing.} System: \textit{``You are a query decomposition and routing expert. Prefer the fewest domains needed and return only valid JSON.''} User prompt: \textit{``You are a query routing agent. Given a user question and a list of available domain experts, decompose the question into sub-questions and route each to the most relevant domain expert(s). If the question is simple and maps to a single domain, return just one sub-question.''} The prompt embeds a summary (label, description, owned topics) for every domain in the org chart.
\item \textbf{Output schema} returned by the LLM:
\begin{quote}\footnotesize\ttfamily
\{"sub\_questions": [\{"question": ...,\\
\ "target\_domains": [...], "context": ...\}]\}
\end{quote}
\end{enumerate}

\textbf{Tools available.}
\begin{itemize}[leftmargin=*,nosep]
\item LLM: \texttt{chat\_completion\_json} for decomposition and routing.
\item Org-chart access for the available-domain summary used in the routing prompt.
\item \texttt{DomainExpertAgent} dispatch (one call per routed domain).
\item Cross-domain context assembly across the answers returned by domain experts before synthesis.
\item \texttt{SynthesisAgent} (Appendix~\ref{app:qa_synthesis}) for final answer assembly.
\end{itemize}

\textbf{Output.} A combined record carrying the original question, the decomposed sub-questions and their routed domains, the per-domain expert responses, the synthesised \texttt{final\_answer} and \texttt{final\_answer\_short}, and overall coverage, confidence, and any remaining knowledge gaps.
\end{agentbox}

\newpage
\subsection{DomainExpertAgent}
\label{app:qa_domain_expert}

\begin{agentbox}
\textbf{Title:} \textit{DomainExpertAgent}\\
\textbf{Role Description.} Answers a sub-question from a single
governed subgraph. Each agent owns one domain, retrieves a
query-focused slice of its subgraph, optionally finds multi-hop paths
or short neighbourhoods around query entities, and produces an
evidence-grounded answer with explicit triple citations.

\textbf{Inputs.}
\begin{itemize}[leftmargin=*,nosep]
\item \texttt{query: str}, optional \texttt{context: str} (routing hints or bridge context from the orchestrator)
\item \texttt{domain: Domain} --- the owned subgraph
\item \texttt{full\_kg: KnowledgeGraph} --- read-only access for multi-hop traversal
\end{itemize}

\textbf{Instruction / Prompt.} Deterministic retrieval followed by a single LLM call.
\begin{enumerate}[leftmargin=*,nosep]
\item \textbf{Query-focused subgraph.} The agent first builds a focused view of its domain that includes the domain label and description, a short summary of prior governance decisions for the domain, and the most query-relevant triples scored by term overlap with the question.
\item \textbf{Multi-hop expansion.} When the question mentions multiple entities present in the graph, the agent additionally retrieves shortest paths between them; when it mentions a single entity, it retrieves a small local neighbourhood of that entity. These paths are appended to the prompt as additional context.
\item \textbf{Answering.} System: \textit{``You are a domain expert for `\{domain.label\}'. Answer queries using ONLY the knowledge graph data provided. Use multi-hop reasoning paths when available to explain indirect connections. Be precise about what you know and don't.''} The user prompt embeds the focused subgraph and any multi-hop paths, the question itself, and a JSON return schema asking for \texttt{answer}, \texttt{coverage}, \texttt{evidence} (a list of cited triples in \texttt{(subject)} \texttt{-[relation]->} \texttt{(object)} form), \texttt{confidence}, and \texttt{out\_of\_scope\_aspects}. The prompt forbids mentioning domain IDs, the routing layer, or the phrase ``knowledge graph,'' and forbids speculation beyond the supplied evidence.
\end{enumerate}

\textbf{Tools available.}
\begin{itemize}[leftmargin=*,nosep]
\item Domain access: \texttt{Domain.} \texttt{get\_subgraph(\ldots)} for owned entities and triples.
\item Graph traversal helpers for shortest paths and entity neighbourhoods over the full graph.
\item LLM: \texttt{chat\_completion\_json} for the single answering call.
\end{itemize}

\textbf{Output.} A per-domain answer record carrying \texttt{answer}, the cited \texttt{evidence} triples, an estimated \texttt{coverage} and \texttt{confidence}, any \texttt{out\_of\_scope\_aspects} the agent could not address, and the originating domain identifier.
\end{agentbox}

\newpage
\subsection{SynthesisAgent}
\label{app:qa_synthesis}

\begin{agentbox}
\textbf{Title:} \textit{SynthesisAgent}\\
\textbf{Role Description.} Combines per-domain expert responses and
any cross-domain bridge context into a final user-facing answer with
both an evidence-grounded prose response and a minimal short-form
answer.

\textbf{Inputs.}
\begin{itemize}[leftmargin=*,nosep]
\item \texttt{question: str}
\item \texttt{domain\_responses: List[Dict]} from each invoked \texttt{DomainExpertAgent}
\item \texttt{cross\_domain\_context: str} --- additional triples that bridge entities mentioned by different experts
\end{itemize}

\textbf{Instruction / Prompt.} Single LLM call. System: \textit{``You are a knowledge synthesis expert. Combine partial answers into a coherent, evidence-grounded response. Return only valid JSON.''} The user prompt lists, for each domain expert, its \texttt{answer}, cited \texttt{evidence}, and \texttt{out\_of\_scope\_aspects}, followed by the bridge triples and the synthesis rules:
\begin{itemize}[leftmargin=*,nosep]
\item \emph{No meta-commentary.} The answer must not mention experts, routing, confidence scores, or the phrase ``knowledge graph''.
\item \emph{Evidence grounding.} Include only claims directly supported by the expert evidence or by triples in the cross-domain context. Both are treated as first-class evidence.
\item \emph{Multi-hop chaining.} For questions that span more than one fact, chain across triples: when an expert supplies a hop-1 entity and the cross-domain context contains a triple about that entity that resolves the next hop, use it rather than abstaining.
\item \emph{Relation paraphrasing.} Treat closely related relation labels as semantic equivalents when the question phrases them differently from the graph (e.g., a partnership relation can support a ``spouse'' or ``collaborator'' question; a ``born in'' relation can support a ``birthplace'' question).
\item \emph{Abstention.} If the supplied evidence does not support an answer, state the limitation briefly rather than speculating.
\item \emph{Two answer forms.} Return both an evidence-grounded \texttt{answer} (1--3 sentences) and a minimal \texttt{short\_answer} (1--5 words; e.g., a name, a place, a year, ``yes'' / ``no'').
\end{itemize}

\textbf{Tools available.}
\begin{itemize}[leftmargin=*,nosep]
\item LLM: \texttt{chat\_completion\_json} for the synthesis call.
\item Cross-domain context assembly used to build the bridge-triple block in the prompt.
\end{itemize}

\textbf{Output.} A JSON record with the prose \texttt{answer}, the minimal \texttt{short\_answer}, an overall \texttt{coverage} and \texttt{confidence}, and a list of remaining \texttt{gaps}. If no domain expert returned anything usable, the orchestrator emits a fixed abstention message instead of calling the LLM.
\end{agentbox}

\newpage
\setcounter{section}{4}
\section{Human Annotation Instructions}
\label{app:anno_ins}
\begin{agentbox}
\footnotesize
Annotation instructions: \\
1. Judge only from the evidence/source text shown. \\
2. Do not use outside knowledge. \\
3. If the triple is plausible but not directly supported, mark partial or no. \\
4. For usefulness, ask whether this triple would help KG search, QA, or reasoning. \\

TRIPLE: (iterative\_deformation) --[Used-for]--> (reconstruction\_process) \\
\\
EVIDENCE / SOURCE TEXT: \\
  iterative deformation of a 3 -- D surface mesh to minimize an objective function
  We propose to incorporate a priori geometric constraints in a 3 -- D stereo reconstruction scheme to cope with the many cases where image information alone is not sufficient to accurately recover 3 -- D shape . Our approach is based on the iterative deformation of a 3 -- D surface mesh to minimize an objective function . We show that combining anisotropic meshing with a non-quadratic approach to regularization enables us to obtain satisfactory reconstruction results using triangulations with few vertices . Structural or numerical constraints can then be added locally to the reconstruction process through a constrained optimization scheme . They improve the reconstruction results and enforce their consistency with a priori knowledge about object shape . The strong description and modeling properties of differential features make them useful tools that can be efficiently used as constraints for 3 -- D reconstruction .\\

Q1. Does the source evidence support the triple? \\
  1. yes \\
  2. partial \\
  3. no \\
  4. unclear \\

Q2. Is the relation/predicate correct? \\
  1. yes \\
  2. partial \\
  3. no \\

Q3. Are the subject and object/endpoints correct? \\
  1. yes \\
  2. partial \\
  3. no

Q4. Is this triple useful for KG search, QA, or graph reasoning? \\
  1. yes \\
  2. maybe \\
  3. no \\

Q5. Did the revision improve the triple? \\
  1. improved \\
  2. same \\
  3. worse \\
  4. unclear \\

Optional short note, or press Enter to skip: \\
\end{agentbox}

\newpage
\end{document}